\documentclass{article}
\usepackage{graphicx,mlspconf}
\usepackage{xcolor}
\usepackage{mathtools}
\usepackage{amssymb}
\usepackage{amsmath}
\usepackage{amsfonts}
\usepackage{appendix}

\newcommand{\argmin}{\mathop{\rm arg~min}\limits}
\usepackage{bm}
\usepackage{subfigure}
\usepackage{arydshln}
\usepackage[ruled,vlined]{algorithm2e}
\SetKwInput{KwInput}{Input}                %
\SetKwInput{KwOutput}{Output} 
\SetKwInput{Preprocessing}{Preprocessing}
\SetKwInput{Initialization}{Initialization}
\renewcommand{\Indentp}[1]{%
  \advance\leftskip by #1
  \advance\skiptext by -#1
  \advance\skiprule by #1}%
\usepackage{multirow}
\usepackage{multicol}
\usepackage{enumitem}
\usepackage{cite}

\usepackage{booktabs}
\usepackage{enumitem}
\setlist[enumerate]{topsep=0.5ex,itemsep=0ex,partopsep=1ex,parsep=0.5ex}
\setlist[itemize]{topsep=0.5ex,itemsep=0ex,partopsep=1ex,parsep=0.5ex}
\setlist[description]{topsep=0.5ex,itemsep=0ex,partopsep=1ex,parsep=0.5ex}

\usepackage{braket}
\usepackage{amsthm}
\theoremstyle{definition}
\newtheorem{assumption}{Assumption}

\renewcommand{\Indp}{\algocf@adjustskipindent\Indentp{\algoskipindent}}
\renewcommand{\Indm}{\algocf@adjustskipindent\Indentp{-\algoskipindent}}
\usepackage{caption}
\copyrightnotice{U.S.\ Government work not protected by U.S.\ copyright}

\copyrightnotice{979-8-3503-2411-2/25/\$31.00 {\copyright}2025 Crown}

\copyrightnotice{979-8-3503-2411-2/25/\$31.00 {\copyright}2025 European Union}

\copyrightnotice{979-8-3503-2411-2/25/\$31.00 {\copyright}2025 IEEE}

\toappear{2025 IEEE International Workshop on Machine Learning for Signal Processing, Aug.\ 31-- Sep.\ 3, 2025, Istanbul, Turkey}

\title{Joint Graph Estimation and Signal Restoration for\\Robust Federated Learning}

\name{Tsutahiro Fukuhara, Junya Hara, Hiroshi Higashi, Yuichi Tanaka}
\address{The University of Osaka}

\begin{document}
\ninept

\maketitle

\begin{abstract}
We propose a robust aggregation method for model parameters in federated learning (FL) under noisy communications. 
FL is a distributed machine learning paradigm in which a central server aggregates local model parameters from multiple clients. 
These parameters are often noisy and/or have missing values during data collection, training, and communication between the clients and server.
This may cause a considerable drop in model accuracy. 
To address this issue, we learn a graph that represents pairwise relationships between model parameters of the clients during aggregation.
We realize it with a joint problem of graph learning and signal (i.e., model parameters) restoration.
The problem is formulated as a difference-of-convex (DC) optimization, which is efficiently solved via a proximal DC algorithm. 
Experimental results on MNIST and CIFAR-10 datasets show that the proposed method outperforms existing approaches by up to $2$--$5\%$ in classification accuracy under biased data distributions and noisy conditions.
\end{abstract}
\begin{keywords}
Federated learning, graph learning, restoration, difference-of-convex.
\end{keywords}

\section{Introduction}
\label{sec:intro}
Federated learning (FL) is an emerging paradigm in machine learning \cite{kairouz2021advances,mcmahan2017communication}. 
It enables collaborative training of neural network(s) by distributing partial models across local servers (referred to as clients). 
These locally trained models are periodically updated by transmitting model parameters from the clients to the central server, followed by aggregation of the parameters and sending them back to the clients.
This scheme could reduce the computational burden on the central server.
FL is considered promising in various applications, such as healthcare, finance, and IoT, to name a few \cite{nguyen2022federated,mammen2021federated}.

FL typically consists of the following three steps\footnote{While there exist various types of FL, we focus on horizontal FL: Hereafter, we refer to it simply as FL.}:
\begin{description}
\item[Initialization] Clients collect different training data. This can be done by distributing different subsets of a training dataset in the central server to all clients, or the clients collect their own data.
\item[Local model update] Each client trains its local model using the collected subset and transmits the trained local parameters to the server.
\item[Global model update]
The server collects and aggregates the local model parameters and obtains global model parameters, which can then be used either as a common model or further personalized by each client. The server may resend the updated global model parameters to all clients for further possible training.
\end{description}
Throughout these three steps, the clients only need to send the model parameters, and their raw data can be kept locally. 
Therefore, it is believed that FL can learn the global model with minimal overhead and also mitigate privacy risks in centralized data collection. 

In practice, the local parameters are often degraded due to data collection, training, and transmission errors to the server.
The degradation occurs due to several factors \cite{ang2020robust,tuor2021overcoming}. 
For example, many communication networks often have low bandwidth, leading to noisy or unstable communications. Similarly, IoT often relies on low-power sensor networks that can be easily disrupted by environmental factors.
This results in noisy and/or missing model parameters.
The aggregation process at the server may be affected by these noises and missing values, which causes severe degradation in accuracy.
In this scenario, robustness against degradation in the global model update is a critical challenge in FL.

The simplest aggregation at the server is computing the mean of the local parameters \cite{mcmahan2016federated}.
While it may alleviate the effect of noise, it could cause oversmoothing since the mean does not consider the differences of the local model parameters, which stem from those of the local training data.
This requires considering appropriate averaging weights based on the underlying relationships among the clients.

Relationships among local model parameters can be mathematically represented by a graph. 
In graph-based FL (GFL) \cite{huang2021personalized,ye2023personalized,ghosh2020efficient}, the server constructs a graph based on similarities among received parameters, i.e., \textit{inter-client graph}, and aggregates these parameters over the graph by graph filtering.
However, the inter-client graph may be noisy if the model parameters contain noise and/or missing values.
Therefore, we face a challenge for a graph-based FL method that takes into account degradation of local model parameters, as well as that in the simple averaging.

In this paper, we propose a graph-based aggregation method for FL under noisy conditions, which simultaneously learns the inter-client graph and calculates aggregated model parameters. Specifically, we perform a joint graph estimation and signal restoration (JGESR) in the global model update phase. 

In the JGESR, we assume that the model parameters vary smoothly on the inter-client graph. Therefore, we jointly optimize the model parameters and inter-client graph so that they satisfy the \textit{graph signal smoothness assumption} \cite{shuman2013emerging}. It is formulated as a non-convex optimization problem, which is generally challenging to solve with finite convergence. To this end, we further reformulate it as an equivalent difference-of-convex (DC) problem and solve it via proximal DC algorithm (PDCA) \cite{gotoh2018dc}. This approach guarantees convergence to at least a local optimum, a desirable property for stable FL applications.

Numerical simulations with two benchmark datasets validate that the proposed method exhibits superior prediction accuracy to existing methods under noisy conditions. 

\textit{Notation}: 
A weighted undirected graph is denoted by $\mathcal{G} = (\mathcal{V},\mathcal{E})$, where $\mathcal{V}$ and $\mathcal{E}$ are sets of nodes and edges.
We use a weighted adjacency matrix $\mathbf{W}$ for representing the connection between the nodes, where its $(m,n)$-element $[\mathbf{W}]_{mn}\geq0$ is the edge weight between the $m$-th and $n$-th nodes; $[\mathbf{W}]_{mn} = 0$ for unconnected nodes. 
The degree matrix $\mathbf{D}$ is a diagonal matrix and its $m$th diagonal element is $[\mathbf{D}]_{mm} = \sum_n[\mathbf{W}]_{mn}$.
The unnormalized graph Laplacian is defined as $\mathbf{L}:=\mathbf{D}-\mathbf{W}$, and the symmetric normalized graph Laplacian is $\widehat{\mathbf{L}}:=\mathbf{D}^{-1/2}\mathbf{L}\mathbf{D}^{-1/2}$.
A graph signal $\mathbf{x}\in\mathbb{R}^K$ is defined as $\mathbf{x} : \mathcal{V}\rightarrow\mathbb{R}^{K}$ where $[\mathbf{x}]_n$ corresponds to the signal value at the $n$th node.
The \textit{Laplacian quadratic form} is defined as $\mathbf{x}^\top\mathbf{L}\mathbf{x}=\sum_{(m,n)\in\mathcal{E}} [\mathbf{W}]_{mn}([\mathbf{x}]_m-[\mathbf{x}]_n)^2$.
This form is often used for a smoothness measure on the graph: A smaller $\mathbf{x}^\top\mathbf{L}\mathbf{x}$ indicates that $\mathbf{x}$ vary more smoothly on $\mathcal{G}$ \cite{shuman2013emerging}.

The symbol $\odot$ indicates Hadamard product and $\operatorname{vec}(\mathbf{A})$ is the vectorization of $\mathbf{A}$. 
We define the $\mathbf{Z}$-norm by $\|\mathbf{A}\|_{\mathbf{Z}}^2 = \operatorname{tr}(\mathbf{A}^\top\mathbf{Z}\mathbf{A})$, where $\mathbf{Z}$ is symmetric and semidefinite. 
The matrix $\mathbf{L} = \operatorname{blkdiag}\bigl(\mathbf{L}_1,\dots,\mathbf{L}_C\bigr)$ denotes the block-diagonal matrix whose $c$th $(c\in\{1,\ldots C\})$ diagonal block is $\mathbf{L}_c$ and whose off-diagonal blocks are all zero.

\begin{figure}[t]
   \centering
  \subfigure[Standard FL]
   {\includegraphics[width=0.495\linewidth]{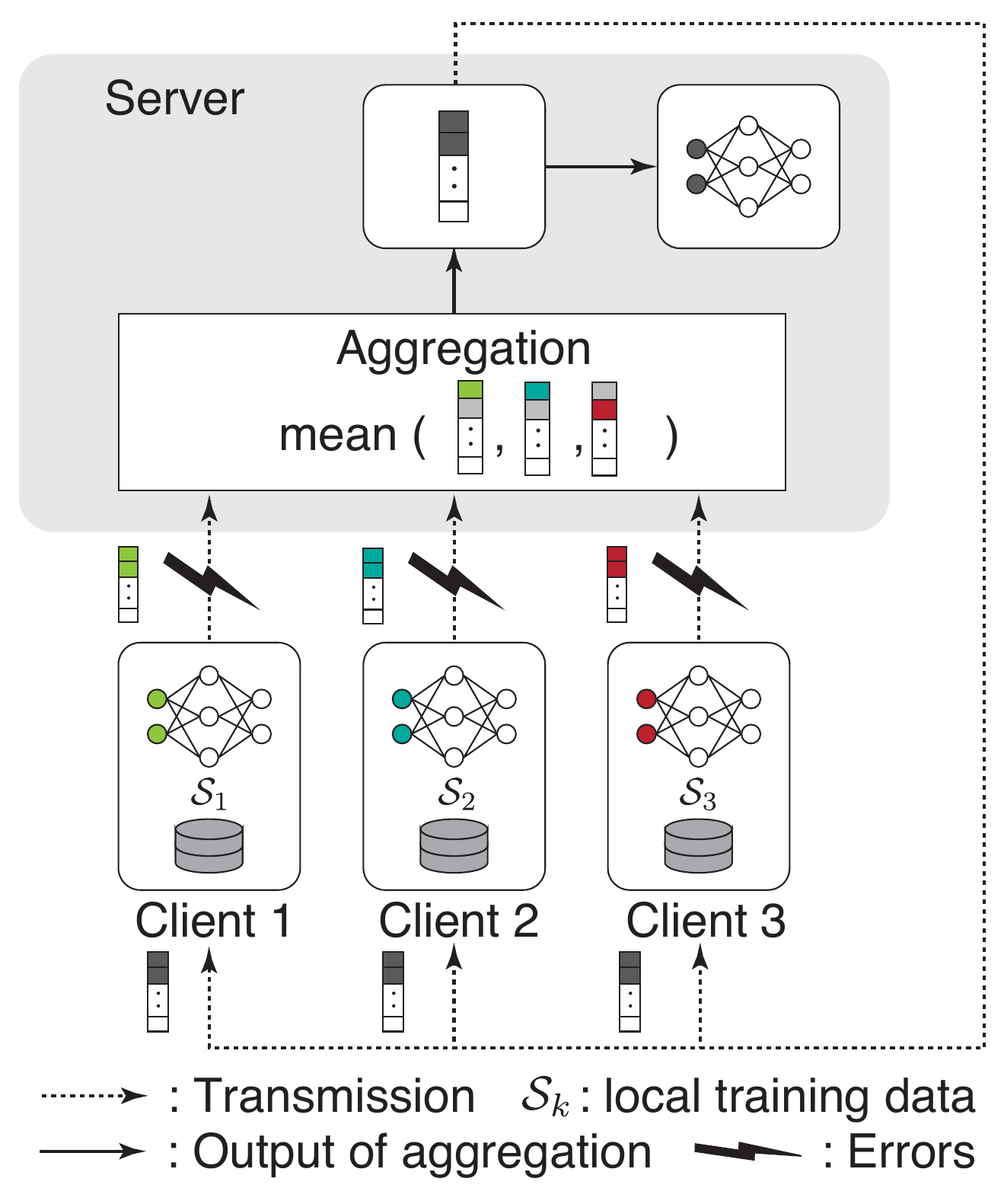}}
  \subfigure[Graph-based FL]
   {\includegraphics[width=0.495\linewidth]{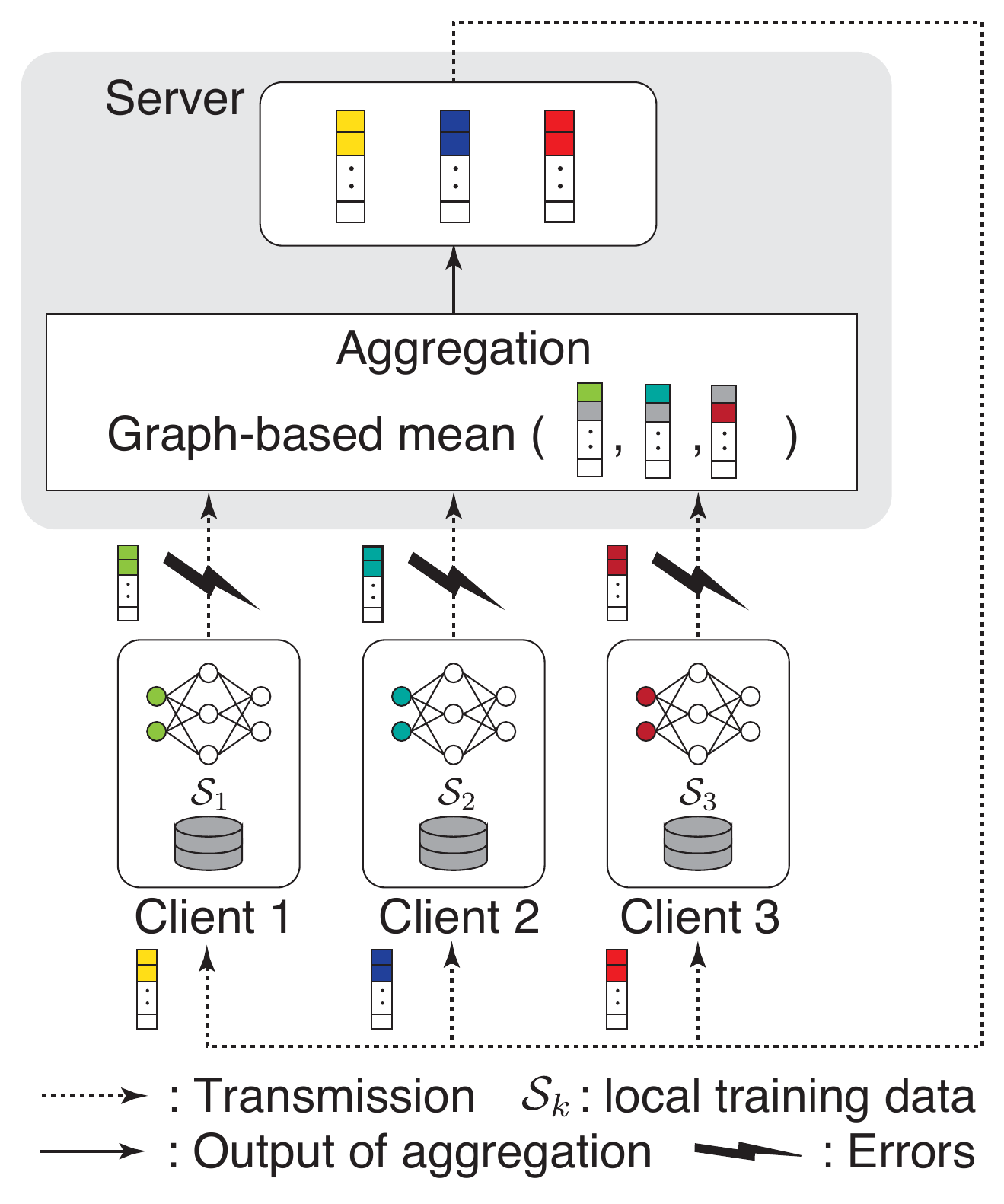}}
 \caption{Overviews of the standard and graph-based FL.}
 \label{fig:overview1}
 \end{figure}
 
\section{Related Works}
In this section, we briefly review the literature on FL.
We first briefly introduce the basic form of FL, then the GFL is reviewed.
We illustrate these FL approaches in Fig.~\ref{fig:overview1}.
\subsection{Standard FL}

In standard FL, the central server aims to make a common global model for all clients by aggregating the local model parameters sent from the clients.

In \cite{mcmahan2016federated}, the server simply aggregates local parameters by computing their mean.
This framework is employed in most existing studies \cite{li2020federated, karimireddy2020scaffold}.
However, these methods tend to oversmooth the model parameters due to their simple aggregation, thereby limiting the accuracy of the global model. Moreover, they are vulnerable to nonuniform dataset sizes and biased data distributions at the clients.

\subsection{Graph-based FL}\label{sec:PFL}

GFL is a branch of FL that aims to adapt the aggregated parameters to be unique for each client.
In other words, it does not have a single global model.
Instead, it individually customizes local model parameters for each client, tailoring them to their local data as well as the global parameters.
GFL uses a graph to represent inter-client relationships, constructed based on the similarities of received parameters.
Then, it aggregates the model parameters over this graph.

Graph-based aggregation can be categorized into cluster-wise and adjacency-wise approaches.
The cluster-wise aggregation 
\cite{ghosh2020efficient,sattler2020clustered} applies clustering to the inter-client graph and calculates intra-cluster averages of the local parameters.
In contrast, the adjacency-wise aggregation \cite{huang2021personalized,ye2023personalized} performs the one-hop graph filtering for parameters at each node.

These methods construct an inter-client graph by calculating pairwise metrics between local parameters.
Unfortunately, it may be corrupted if model parameters themselves contain noise and/or missing values. Consequently, the aggregation qualities highly depend on the accuracy of the inter-client graph.

\section{Preliminaries}\label{sec:preliminary}

In this section, we introduce the building blocks used in the proposed FL.
First, we formalize the GFL framework. 
Second, we introduce a graph-based aggregation method based on the graph signal smoothness assumption.
Third, we describe a graph learning method.

\subsection{Framework of Graph-based Federated Learning}\label{ssec:framework_pfl}

Let us consider GFL with $K$ clients, where the $k$th client ($k\in\{1,\ldots,K\}$) has a local training dataset $\mathcal{S}_k$ which does not overlap with the other local datasets. 
Let $\mathbf{x}_k$ denote the model parameters of the $k$th client and $f_k(\mathbf{x}_k)$ be its local loss function. 
The goal of GFL is to minimize the weighted sum of individual loss functions designed for each client. 
By utilizing matrix form $\mathbf{X}=[\mathbf{x}_1,\ldots,\mathbf{x}_K]^\top\in\mathbb{R}^{K\times d}$, this can be formulated as follows:
\begin{equation} \label{eq:obj1}
    \min_{\mathbf{X}} \; \mathcal{F}(\mathbf{X})+R(\mathbf{X}), 
\end{equation}
where $\mathcal{F}(\mathbf{X})=\sum_{k=1}^K\zeta_k f(\mathbf{x}_k)$,
$\zeta_k = \frac{|\mathcal{S}_k|}{\sum^K_{k=1}|\mathcal{S}_k|}$ is a weight for the $k$th client reflecting its proportion to the total data size, and $R(\cdot)$ is a properly designed regularization function.

Since directly solving \eqref{eq:obj1} is challenging, in general, GFL introduces auxiliary variables of $\mathbf{X}$ as $\mathbf{\Psi} = [\bm{\psi}_1,\ldots,\bm{\psi}_K]^\top\in\mathbb{R}^{K\times d}$.
Then, \eqref{eq:obj1} is reformulated as
\begin{equation}\label{eq:obj2}
\min_{\mathbf{X},\bm{\Psi}} \mathcal{F}(\mathbf{X})+\frac{\mu}{2}\|\mathbf{X}-\bm{\Psi}\|_{\mathbf{Z}}^2+R(\mathbf{\Psi}),
\end{equation}
where $\mathbf{Z} = \operatorname{diag}(\zeta_1,\ldots,\zeta_K)$ and $\mu$ is a parameter.
The second term in \eqref{eq:obj2} encourages the fidelity between primal and auxiliary variables. 
To solve \eqref{eq:obj2}, GFL typically involves the following steps: 
\begin{description}
    \item[Initialization] 
    Let the superscript $\cdot^{(r)}$ indicate the round of updates.
    The server distributes the global model parameters to each client $k\in\{1,\ldots,K\}$:
    \begin{equation*}
        \mathbf{\mathbf{x}}_{k,\text{global}}^{(r)} = \bm{\psi}_k^{(r)}. 
    \end{equation*}
    The initial model parameters $\bm{\psi}_k^{(0)}$ are typically set randomly.
    \item[Local model update] Each client updates its local model by minimizing \eqref{eq:obj2} as follows. 
    \begin{equation}\label{eq:client}
        \mathbf{x}_{k,\text{local}}^{(r+1)}= \argmin_{\mathbf{x}_k}\,f_k(\mathbf{x}_k)+\frac{\mu}{2}||\mathbf{x}_k-\mathbf{\mathbf{x}}_{k,\text{global}}^{(r)}||_2^2.
    \end{equation}
    \item[Global model update] The server aggregates the locally updated parameters by minimizing \eqref{eq:obj2} w.r.t. $\mathbf{\Psi}$, i.e.,
    \begin{equation}\label{eq:server}
        \bm{\Psi}^{(r+1)}=\argmin_{\mathbf{\Psi}}\;\frac{\mu}{2}\|\mathbf{\Psi}-\mathbf{X}_{\text{local}}^{(r+1)}\|^2_{\mathbf{Z}}+R(\mathbf{\Psi}),
    \end{equation}
        where $\mathbf{X}_{\text{local}}^{(r+1)}=[\mathbf{x}_{1,\text{local}}^{(r+1)},\ldots,\mathbf{x}_{K,\text{local}}^{(r+1)}]^\top$ are the parameters received by the server at the current round.
\end{description}
As observed, GFL is an alternative optimization approach: The clients update each of $\{\mathbf{x}_{k,\text{local}}\}_{k=1}^K$ in \eqref{eq:obj2} and the server updates $\{\bm{\psi}_k\}_{k=1}^K$.  We illustrate the GFL approach in Fig.~\ref{fig:overview1}(b).

Note that the design of $R(\cdot)$ in \eqref{eq:server} plays a key role in GFL.
For example, \cite{arivazhagan2019federated} enforce that specific parameters share the same values across all clients.
A self‑regularization term to avoid overfitting is proposed in \cite{smith2017federated}.

\subsection{Graph-based Aggregation}\label{ssec:graph_aggregate}
Here, we focus on the details of the global model update introduced in the previous subsection \cite{huang2021personalized,ye2023personalized,sattler2020clustered}.

To begin with, we assume the following condition:
\begin{assumption}[Smoothness assumption \cite{shuman2013emerging}]\label{assump:smoothness}
    Suppose that a connected graph $\mathcal{G}=(\mathcal{V},\mathcal{E})$ with $|\mathcal{V}|=K$ is given and the corresponding graph Laplacian is denoted by $\mathbf{L}$. 
    Here, we assume that $\bm{\psi}_k$ is a graph signal on $\mathcal{G}$.
    If $\{\bm{\psi}_k\}_{k=1}^K$ vary smoothly over $\mathcal{G}$, the feasible set of signals is given by
    \begin{equation}
        \mathcal{T}_{\text{SM}}\coloneqq\left\{\mathbf{\Psi}\in\mathbb{R}^{K\times d}\mid \operatorname{tr}(\mathbf{\Psi}^\top\mathbf{L}\mathbf{\Psi})\leq \delta\right\},
    \end{equation}
    where $\delta>0$ is a small constant.
\end{assumption}
Under Assumption~\ref{assump:smoothness}, one incorporates the Laplacian quadratic form $\operatorname{tr}(\mathbf{\Psi}^\top\mathbf{L}\mathbf{\Psi})$ into \eqref{eq:server} as
\begin{equation}\label{eq:special_case1}
\begin{split}
    \bm{\Psi}^{(r+1)} &=\argmin_{\mathbf{\Psi}}\;\frac{\mu}{2}||\mathbf{\Psi}-\mathbf{X}^{(r+1)}_{\text{local}}||^2_{\mathbf{Z}} + \alpha\,\text{tr}({\mathbf{\Psi}}^\top\mathbf{L}\mathbf{\Psi})\\
    &=\left(\mathbf{Z}+\frac{2\alpha}{\mu}\mathbf{L}\right)^{-1}\mathbf{Z}\mathbf{X}^{(r+1)}_{\text{local}}
\end{split}
\end{equation}
where $\alpha>0$ is a parameter.
Intuitively, \eqref{eq:special_case1} can be viewed as graph low-pass filtering of weighted row entries in $\mathbf{X}^{(r+1)}_{\text{local}}$ based on the smoothness assumption. 

In fact, \eqref{eq:special_case1} is a generic form of existing graph-based aggregation methods, and this fact is illustrated below:
\begin{description}
\item[Cluster‑wise aggregation] 
        It partitions the graph into clusters and averages the received model parameters among the nodes within each cluster \cite{sattler2020clustered}. 
        This approach is a special case of \eqref{eq:special_case1} when the graph $\mathcal{G}$ consists of $C$ disjoint subgraphs $\mathcal{G}_c=(\mathcal{V}_c,\mathcal{E}_c)$ with $|\mathcal{V}_c|=K_c$ for $c=1,\ldots,C$, and the data distribution across clients is uniform, i.e., $\mathbf{Z}=\mathbf{I}$. Under these conditions, $\mathbf{L} = \operatorname{blkdiag}(\mathbf{L}_1,\dots,\mathbf{L}_C)$. If the parameter $\alpha$ in \eqref{eq:special_case1} is sufficiently large, \eqref{eq:special_case1} simplifies to
        \begin{equation}\label{eq:special_case2}
        \begin{split}
            &\mathbf{\Psi}^{(r+1)}\\
            &= \lim_{\alpha\rightarrow\infty}\left(\mathbf{I}+\frac{2\alpha}{\mu}\operatorname{blkdiag}(\mathbf{L}_1,\ldots,\mathbf{L}_C)\right)^{-1}\mathbf{X}^{(r+1)}_{\text{local}}\\
            &=\lim_{\alpha\rightarrow\infty}\sum_{c\in\mathcal{C}}\mathbf{U}_c{\textstyle\operatorname{diag}\left(1,\ldots,\frac{\mu}{\mu+2\alpha\lambda_{K_c}}\right)}\mathbf{U}_c^\top\mathbf{X}^{(r+1)}_{\text{local}}\\
            &= \sum_{c\in\mathcal{C}}\frac{1}{K_c}\mathbf{1}_{\mathcal{V}_c}\mathbf{1}_{\mathcal{V}_c}^\top\mathbf{X}^{(r+1)}_{\text{local}},
        \end{split}
        \end{equation}
        where $\mathcal{C}=\{1,\ldots,C\}$, $[\mathbf{1}_{\mathcal{V}_c}]_i=1$ if $i\in\mathcal{V}_c$ and $[\mathbf{1}_{\mathcal{V}_c}]_i=0$ otherwise, and $\mathbf{U}_c\in\mathbb{R}^{K\times K_c}$ is the submatrix of $\mathbf{U}$ collecting $K_c$ columns that corresponds to the subgraph $\mathcal{G}_c$. In the third equality in \eqref{eq:special_case2}, we use $\mathbf{U}_c\operatorname{diag}(1,0,0,\ldots)\mathbf{U}_c=\frac{1}{K_c}\mathbf{1}_{\mathcal{V}_c}\mathbf{1}_{\mathcal{V}_c}^\top$. 
        This is nothing but the cluster-wise average of local model parameters.
\item[Adjacency‑wise aggregation]
        It updates parameters on each node by averaging those on its one-hop neighbors. Suppose that the normalized Laplacian is used as $\mathbf{L}$. When $\alpha = 2\mu$ and $\mathbf{Z}=\mathbf{I}$ in \eqref{eq:special_case1}, the solution can be approximated by
        \begin{equation}\label{eq:special_case3}
        \begin{split}
            \mathbf{\Psi}^{(r+1)}
            &= (\mathbf{I}+\widehat{\mathbf{L}})^{-1}\mathbf{X}^{(r+1)}_{\text{local}}\\
            &\approx (\mathbf{I}-\widehat{\mathbf{L}})\mathbf{X}^{(r+1)}_{\text{local}}= \mathbf{D}^{-1/2}\mathbf{W}\mathbf{D}^{-1/2}\mathbf{X}^{(r+1)}_{\text{local}},
        \end{split}
        \end{equation} 
        where we utilize the first-order Taylor expansion from the first line to the second one.
\end{description}
From these examples, we can see that \eqref{eq:special_case1} is a generalized version of the existing graph-based aggregation methods. We later use the form in \eqref{eq:special_case1} for our formulation.

Throughout this subsection, we assume the inter-client graph is given a priori.
However, this is not the case in GFL. Therefore, one needs to estimate the inter-client graph from given signals prior to performing the graph-based aggregation. 

\begin{figure*}[h]
    \centering
    \includegraphics[width = 0.9\linewidth]{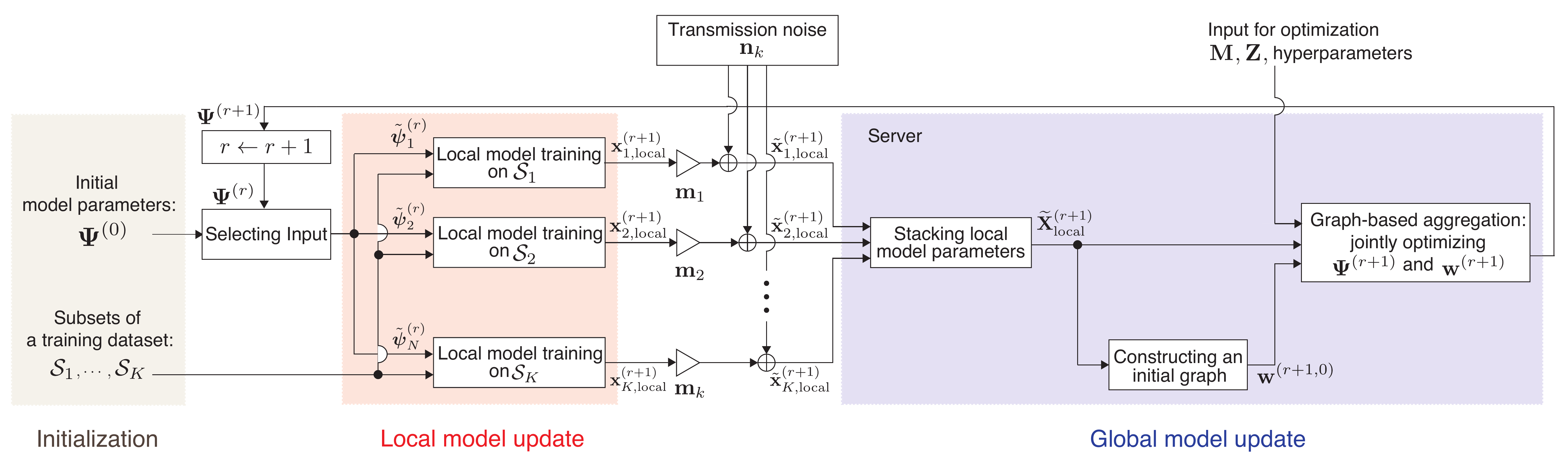}

    \caption{Overview of the proposed FL.}
    \label{fig:proposed}
\end{figure*}

\subsection{Graph Learning}\label{ssec:graph_estimate}

We introduce a graph learning method based on the smoothness assumption in Assumption~\ref{assump:smoothness}. One representative formulation is given by \cite{kalofolias2016learn}:
\begin{equation}\label{eq:graph_estimate_2}
    \mathbf{W}=\argmin_{\mathbf{W}\in\mathcal{W}} \alpha\operatorname{tr}(\mathbf{W}\mathcal{D}(\mathbf{X}))-\beta\bm{1}^\top\log(\mathbf{W}\bm{1})+\gamma\|\mathbf{W}\|_1,
\end{equation}
where $\mathcal{W}=\{\mathbf{W}\in\mathbb{R}^{K\times K}\mid \mathbf{W}=\mathbf{W}^\top,[\mathbf{W}]_{m\neq n}\geq 0,[\mathbf{W}]_{mm}=0\}$ and $\mathcal{D}(\mathbf{X})$ is the Euclidean distance matrix for row entries in $\mathbf{X}$, i.e., $[\mathcal{D}(\mathbf{X})]_{mn}=\sum_{m,n}\|\mathbf{x}_m-\mathbf{x}_n\|^2$. 
In \eqref{eq:graph_estimate_2}, the first term is equivalent to the Laplacian quadratic form: $\operatorname{tr}(\mathbf{W}\mathcal{D}(\mathbf{X})) = \operatorname{tr}(\mathbf{X}^\top\mathbf{L}\mathbf{X})$, which enforces the smoothness of $\mathbf{X}^{(r)}$ on the graph $\mathcal{G}$.
The second term (with the parameter $\beta>0$) penalizes nodes so that they are not isolated, and the third term (with the parameter $\gamma>0$) encourages the sparsity of edges. 
Since \eqref{eq:graph_estimate_2} is convex, it is efficiently solved by convex algorithms \cite{boyd2004convex}.

While the graph learned with \eqref{eq:graph_estimate_2} is robust to noise, the graph learning followed by signal aggregation may result in a suboptimal solution.

\section{Joint Graph Estimation and Signal Restoration for FL}

In this section, we present a graph-based aggregation method for GFL based on JGESR.
We first formulate a graph-based aggregation in the global model update as a DC optimization problem and then solve it using PDCA. We illustrate the overview of the proposed FL in Fig.~\ref{fig:proposed}.

\subsection{Problem Formulation}

Many FL studies assume that model parameters are degraded by noise and/or missing values during the transmission from clients to the server \cite{amiri2020federated, yang2020federated} (see \eqref{eq:server}).
This scenario arises from the fact that clients often need to transmit parameters under resource constraints, such as limited transmit power and bandwidth. Therefore, in general, the parameters $\mathbf{x}_{k,\text{local}}^{(r+1)}$ in \eqref{eq:server} can be modeled as \cite{amiri2020federated,yang2020federated}
\begin{equation}
        \label{eq:noise1}
        \tilde{\mathbf{x}}_{k,\text{local}}^{(r+1)} = \mathbf{m}_{k}\odot\mathbf{x}_{k,\text{local}}^{(r+1)}+\mathbf{n}_k,\quad\mathbf{n}_k\sim\mathcal{N}(\mathbf{0},\sigma_k^2\mathbf{I}),
    \end{equation}
    where $\mathbf{m}_k\in\mathbb{R}^{d}$ is a vector representing the partial loss or gain changes, and $\mathbf{n}_k\in\mathbb{R}^{d}$ is additive white Gaussian noise with standard deviation $\sigma_k$.

First, we simply merge \eqref{eq:special_case1}, \eqref{eq:graph_estimate_2}, and \eqref{eq:noise1} into a single optimization problem:
\begin{equation}\label{eq:ser_obj1}
\begin{split}
    \mathbf{\Psi}^{(r+1)}
    &=\argmin_{\mathbf{\Psi},\mathbf{W}} \;\frac{\mu}{2}||\mathbf{M}\odot\mathbf{\Psi}-\widetilde{\mathbf{X}}^{(r+1)}_{\text{local}}||^2_{\mathbf{Z}}+\alpha \operatorname{tr}(\mathbf{W}\mathcal{D}(\mathbf{\Psi}))\\
    &\quad -\beta\mathbf{1}^\top\log{(\mathbf{W}\mathbf{1})}+ \gamma||\mathbf{W}||_1 +\iota_{\mathcal{W}}(\mathbf{W}),
\end{split}
\end{equation}
where $\mathbf{M}=[\mathbf{m}_1,\ldots,\mathbf{m}_K]^\top\in\mathbb{R}^{K\times d}$, $\widetilde{\mathbf{X}}^{(r+1)}_{\text{local}} = [\tilde{\mathbf{x}}^{(r+1)}_{1,\text{local}},\ldots,\\\tilde{\mathbf{x}}^{(r+1)}_{K,\text{local}}]^\top$, and 
$\iota_{\mathcal{W}}$ is the indicator function over the set $\mathcal{W}$, i.e.,
\begin{equation}\label{eq:indicator}
    \iota_{\mathcal{W}}(\mathbf{W}) =
    \begin{cases} 
        0 & \text{if} \;\mathbf{W}\in\mathcal{W},\\
        +\infty & \text{otherwise}.
    \end{cases}
\end{equation}
Note that \eqref{eq:ser_obj1} is non-convex because $\operatorname{tr}(\mathbf{W}\mathcal{D}(\mathbf{\Psi}))$ is biconvex in both $\mathbf{W}$ and $\mathbf{\Psi}$, which is difficult to solve with finite convergence.

While one might consider separately optimizing $\mathbf{W}$ and $\mathbf{\Psi}$ in \eqref{eq:ser_obj1} via alternating minimization, it risks convergence to poor local minima due to the biconvex term.
To address this issue, we reformulate \eqref{eq:ser_obj1} as a difference-of-convex (DC) optimization problem, which allows us to establish local convergence guarantees \cite{gotoh2018dc}.

Here, we let $\mathbf{w}=\text{upper}(\mathbf{W})\in\mathbb{R}^{K(K-1)/2}$ is the vector form representation of upper triangular entries of $\mathbf{W}$, and 
$\mathbf{T}\in\mathbb{R}^{K(K-1)/2\times K^2}$ is the transform satisfying $\mathbf{T}\text{vec}(\mathcal{D}(\mathbf{\Psi})) = \text{upper}(\mathcal{D}(\mathbf{\Psi}))$.
Then, we can transform $\operatorname{tr}(\mathbf{W}\mathcal{D}(\mathbf{\Psi}))$ in \eqref{eq:ser_obj1} as:
\begin{equation} \label{eq:biconvex1}
\begin{split}
    &\operatorname{tr}(\mathbf{W}\mathcal{D}(\mathbf{\Psi}))\\
     & = \frac{1}{2}||\mathbf{W}+\mathcal{D}(\mathbf{\Psi})||^2_F  -\frac{1}{2}(||\mathbf{W}||^2_F+||\mathcal{D}(\mathbf{\Psi})||^2_F)\\
    &=||\mathbf{w}+\mathbf{T}\operatorname{vec}(\mathcal{D}(\mathbf{\Psi}))||^2_2-(||\mathbf{w}||^2_2+||\mathbf{T}\operatorname{vec}(\mathcal{D}(\mathbf{\Psi}))||^2_2).
\end{split}
\end{equation}
Note that, since both terms in \eqref{eq:biconvex1} are convex in terms of $\mathbf{w}$ and $\mathbf{\Psi}$,
it is a DC decomposition of the biconvex terms in \eqref{eq:ser_obj1}.
Consequently, we can organize a DC form of \eqref{eq:ser_obj1} as

\begin{equation}\label{eq:ser_obj3}
\begin{split}
    \argmin_{\mathbf{\Psi},\mathbf{w}}& \;f(\mathbf{\Psi},\mathbf{w})+g(\mathbf{\Psi},\mathbf{w})-h(\mathbf{\Psi},\mathbf{w}),
\end{split}
\end{equation}
where 
\begin{subequations}
\begin{align}
     f(\mathbf{\Psi},\mathbf{w}) & = \frac{\mu}{2}||\mathbf{M}\odot\mathbf{\Psi}-\widetilde{\mathbf{X}}^{(r+1)}_{\text{local}}||^2_\mathbf{Z}+\alpha\|\mathbf{w}+\mathbf{T}\text{vec}(\mathcal{D}(\mathbf{\Psi}))||^2_2,\\
    g(\mathbf{\Psi},\mathbf{w})& = -\beta\mathbf{1}^\top\log{(\mathbf{B}\mathbf{w})}+ \gamma||\mathbf{w}||_1+\iota_{\geq0}(\mathbf{w})\label{eq:g()},\\
    h(\mathbf{\Psi},\mathbf{w})& = \alpha\bigl(||\mathbf{w}||_2^2+||\mathbf{T}\text{vec}(\mathcal{D}(\mathbf{\Psi}))||_2^2\bigr),
\end{align}
\end{subequations}
in which 
$\mathbf{B}\in\mathbb{R}^{K\times K(K-1)/2}$ is the transform satisfying $\mathbf{W}\mathbf{1}=\mathbf{B}\mathbf{w}$, and $\iota_{\geq 0}(\mathbf{w})$ enforces the non-negativity on the entries of $\mathbf{w}$ (see \eqref{eq:indicator}).

\subsection{Proposed Algorithm}

We use PDCA \cite{gotoh2018dc} to solve \eqref{eq:ser_obj3} because it provides finite iteration convergence guarantees for DC problems under mild conditions\footnote{Our method satisfies the conditions: We omit their details due to space limitation.}, and it can naturally handle the differentiable but non-smooth functions like those in \eqref{eq:g()}. 

Algorithm \ref{alg:outline} presents the proposed GFL integrating PDCA for solving \eqref{eq:ser_obj3}.
While we omit the details due to the limitation of space, in the algorithm, we use the following operators:
\begin{itemize}
    \item $\nabla_{\bm{\Psi}}f(\mathbf{\Psi},\mathbf{w})= \mu\mathbf{M}\odot(\mathbf{Z}(\mathbf{M}\odot(\mathbf{\Psi}-\widetilde{\mathbf{X}}^{(r+1)}_{\text{local}})))$
    \item[]$\;\qquad\qquad\quad\,\,\,+\alpha\mathcal{D}^*(\text{vec}^{-1}(\mathbf{T}^\top
    (\mathbf{w}+\mathbf{T}\text{vec}(\mathcal{D}(\mathbf{\Psi})))))\mathbf{\Psi}$, 
    \item $\nabla_{\mathbf{w}}f(\mathbf{\Psi},\mathbf{w})=\alpha(\mathbf{w}+\mathbf{T}\text{vec}(\mathcal{D}(\mathbf{\Psi})))$,
        \item  $\nabla_{\mathbf{\Psi}} h(\mathbf{\Psi},\mathbf{w})= \alpha\mathcal{D}^*(\text{vec}^{-1}(\mathbf{T}^\top\mathbf{T}\text{vec}(\mathcal{D}(\mathbf{\Psi}))))\mathbf{\Psi}$,
         \item $\nabla_{\mathbf{w}}h(\mathbf{\Psi},\mathbf{w}) = \alpha\mathbf{w}$, and
    \item $\operatorname{prox}_{g/\rho}(\mathbf{\Theta})\coloneqq\argmin_{\mathbf{\Phi}}g(\mathbf{\Theta})+\frac{\rho}{2}\|\mathbf{\Phi}-\mathbf{\Theta}\|_F^2$, 
\end{itemize}
where $\mathcal{D}^*(\mathbf{H}) = \text{diag}(\mathbf{H1})+\text{diag}(\mathbf{H}^\top\mathbf{1})-2\mathbf{H}$ is the adjoint operator of $\mathcal{D}$
, and $\text{vec}^{-1}(\cdot)$ is the inverse of $\text{vec}(\cdot)$. 
Since \eqref{eq:g()} consists of multiple convex functions, we can easily solve $\operatorname{prox}_{g/\rho}$ in Algorithm~\ref{alg:outline} by using existing convex algorithms \cite{boyd2004convex}.

\begin{algorithm}[h]
\DontPrintSemicolon
\setlength{\abovedisplayskip}{0pt}
\setlength{\belowdisplayskip}{0pt}
\KwInput{$\mathbf{M}$, $\mathbf{Z}$, $R$, $E$, $\alpha,\beta,\gamma, \epsilon,\eta,\mu,\rho$.}
\Initialization{Set local model weights $\bm{\psi}_k^{(0)}$ randomly.}
\For{$r = 0,1,\ldots,R-1$}{
 \textit{\# Local model update} \\
    \For{\emph{all $k\in\{1,\ldots,K\}$ in parallel}}
        {
        $\mathbf{x}_{k,\text{local}}^{(r+1,0)}=\bm{\psi}_k^{(r)}, \mathbf{x}_{k,\text{global}}^{(r)}=\bm{\psi}_k^{(r)}$\\
        \For{$e=0,1,\ldots,E-1$}{
        $\mathbf{x}_{k,\text{local}}^{(r+1,e+1)}=\mathbf{x}_{k,\text{local}}^{(r+1,e)}-\eta \left( \nabla f_k(\mathbf{x}_{k,\text{local}}^{(r+1,e)}) + \mu (\mathbf{x}_{k,\text{local}}^{(r+1,e)} - \mathbf{x}^{(r)}_{k,\text{global}}) \right)$
        
            }
        Send updated model $\mathbf{x}^{(r+1,E+1)}_{k,\text{local}}$ to the server.
        }
    \textit{\# Global model update (JGESR via PDCA)}\\
    Calculate $\mathbf{w}^{(r+1,0)}$ using $\widetilde{\mathbf{X}}^{(r+1)}_{\text{local}}$.\\
    $\mathbf{\Theta}=(\mathbf{\Psi}^{(r+1,0)} = \widetilde{\mathbf{X}}^{(r+1)}_{\text{local}},\mathbf{w}^{(r+1,0)})$\\
    \For{$t = 0,1,\ldots$}{
      $\mathbf{s}^{(t+1)} = (\nabla_{\mathbf{\Psi}}h(\mathbf{\Theta}),\nabla_{\mathbf{w}}h(\mathbf{\Theta}))$\\
      $\mathbf{u}^{(t+1)} = (\nabla_{\mathbf{\Psi}}f(\mathbf{\Theta}),\nabla_{\mathbf{w}}f(\mathbf{\Theta}))$\\
     $\mathbf{\Theta}=\text{prox}_{g/\rho}(\mathbf{\Theta}-\frac{1}{\rho}(\mathbf{u}^{(t+1)}-\mathbf{s}^{(t+1)}))$\\
     Check the convergence criterion:\\
     \If{$||\mathbf{\Psi}^{(r+1,t+1)}-\mathbf{\Psi}^{(r+1,t)}||_F<\epsilon$}{\textbf{break}}
    Send the aggregated model $\bm{\psi}_k^{(r+1)}$ for each client.\\
    }
}
\KwOutput{$\mathbf{\Psi}^{(R)}$}
\caption{Proposed GFL}\label{alg:outline}
\end{algorithm}

\section{Experiments}
In this section, we evaluate the performance of the proposed method for image classification using two benchmark datasets.

\subsection{Experimental setup}
\textit{Datasets}: 
In the experiments, we use the following two benchmark datasets: 
\begin{description}
    \item[MNIST \cite{lecun1998gradient}:] This dataset consists of 70,000 (60,000 for training and 10,000 for test) handwritten digit images ranging from 0 to 9.
    \item[CIFAR-10 \cite{krizhevsky2012imagenet}:] It consists of 60,000 color images (50,000 for training and 10,000 for test) across 10 distinct classes.
\end{description} 
We partition the training data of each dataset into local training datasets according to a Dirichlet distribution $\operatorname{Dir}_K(\kappa)$ \cite{mcmahan2016federated}, where $\kappa$ is a parameter controlling the bias in the data distribution\footnote{In a Dirichlet distribution $\operatorname{Dir}_K(\kappa)$, a smaller value of $\kappa$ leads to a stronger bias in the data distribution.}.
To simulate biased distributions, we set $\kappa=0.05$.
The local training datasets are divided into $75\%$ for local training and $25\%$ for local test.

\noindent
\textit{Comparison methods and performance measure}:
We compare our method with five existing FL methods: 
\begin{description}
    \item[Standard FL:]FedAvg \cite{mcmahan2016federated}, FedProx \cite{li2020federated}.
    \item[Graph-based FL:]CFL \cite{sattler2020clustered}, FedAMP \cite{huang2021personalized}, and pFedGraph \cite{ye2023personalized}.
\end{description}
In addition, we also compare our method with a simpler baseline (Two-step) for solving \eqref{eq:ser_obj1}, which separately optimizes $\mathbf{W}$ and $\mathbf{\Psi}$ via alternating minimization \cite{dong2016learning}.
In comparison with alternative methods, we set $\mathbf{m}_k = \mathbf{1}$ in \eqref{eq:noise1}, i.e., no transmission errors, for a fair comparison since the other FL methods do not consider transmission error.
The standard deviations in \eqref{eq:noise1} are set proportional to the average absolute value of the initial model parameters: 
$\sigma_{k} = s\,\mathbb{E}_{k}[|\bm{\psi}_k^{(0)}|]$, where $s$ is a scaling factor. 
We use the same initial model parameters $\bm{\psi}_k^{(0)}$ for all methods. 

After the comparison with the alternative methods, we show experimental results to validate the robustness of the proposed method against partial parameter loss.
In this case, we introduce random binary masks $\mathbf{m}_k\in\{0,1\}^d$ and measure the performance under various missing rates.

We perform $10$ independent runs of image classification on both datasets.
Then, we measure the local test accuracy for each client in each run and evaluate the average across all clients and runs.

\noindent
\textit{Training and optimization configuration}: 
We set $K=20$ clients and perform $R = 30$ communication rounds.  
In the local training, all clients use a simple CNN with two convolutional layers and two fully-connected layers. We employ SGD as the optimizer \cite{mcmahan2016federated}. 
The local epochs and step size are set to $E = 5$ and $\eta = 0.01$, respectively. 

In Algorithm~\ref{alg:outline}, we calculate $\mathbf{w}^{c(r+1,0)}$ using the cosine similarity among the row entries of $\tilde{\mathbf{X}}^{(r+1)}_{\text{local}}$, and experimentally set the hyperparameters in JGESR to $\alpha = 0.05$, $\mu=\beta=\gamma=\rho=1.0$, $\epsilon = 0.001$ (see Algorithm~\ref{alg:outline}).

\begin{table}[t]
\centering
\caption{Average accuracy comparison $(\%)$ on two datasets with different noise levels $s$.}
\begin{tabular}{lccccc}
\toprule
Dataset & \multicolumn{2}{c}{MNIST} & \multicolumn{2}{c}{CIFAR-10} \\
\cmidrule(lr){2-3} \cmidrule(lr){4-5}
Noise level $s$
& $0.1$ & $0.2$ & $0.1$ & $0.2$ \\
\midrule
Proposed (PDCA) & \textbf{95.17} & \textbf{92.68} & \textbf{83.86} & \textbf{78.04} \\
Proposed (Two-step) & 88.99 & 83.77 & 72.30 & 69.90 \\ 
\hdashline
FedAvg      & 90.48 & 89.96 & 77.31 & 74.80 \\
FedProx     & 90.38 & 91.19 & 76.69 & 74.10 \\
CFL         & 92.96 & 91.21 & 78.35 & 75.70 \\
FedAMP      & 93.12 & 92.40 & 78.70 & 75.12 \\
pFedGraph   & 92.76 & 92.01 & 78.19 & 75.24 \\
\bottomrule
\end{tabular}
\label{tab:result}
\end{table}

\begin{figure}
    \centering
    \includegraphics[width=0.7\linewidth]{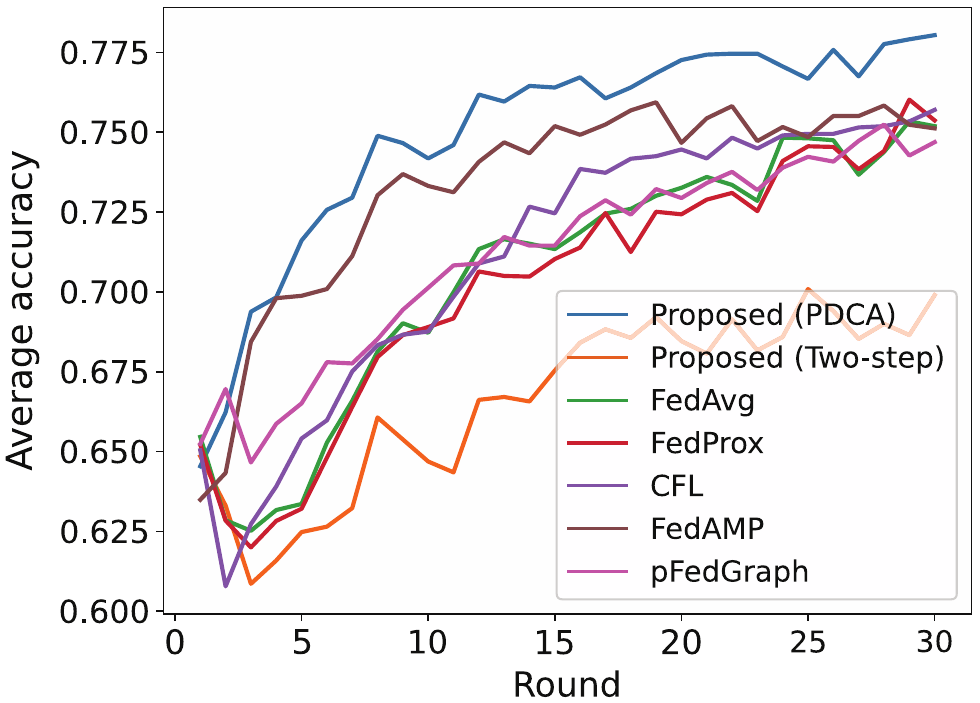}
    \caption{Average accuracies at each communication round on CIFAR-10 under a noise level $s=0.2$.}
    \label{fig:plot_round}
\end{figure}

\begin{figure}[t]
   \centering
  \subfigure[MNIST]
   {\includegraphics[width=0.495\linewidth]{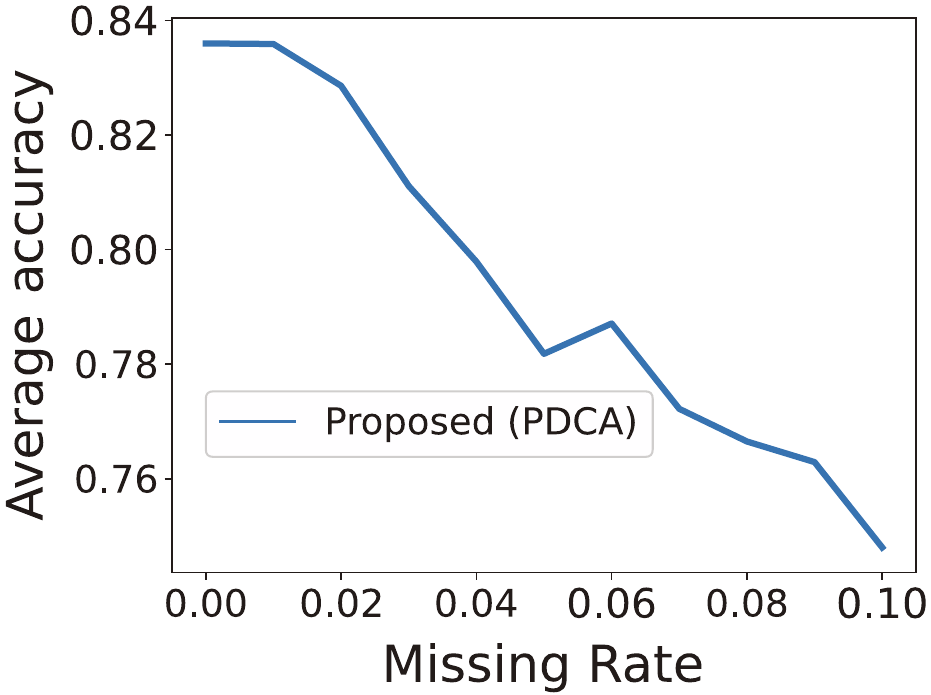}}
  \subfigure[CIFAR-10]
   {\includegraphics[width=0.495\linewidth]{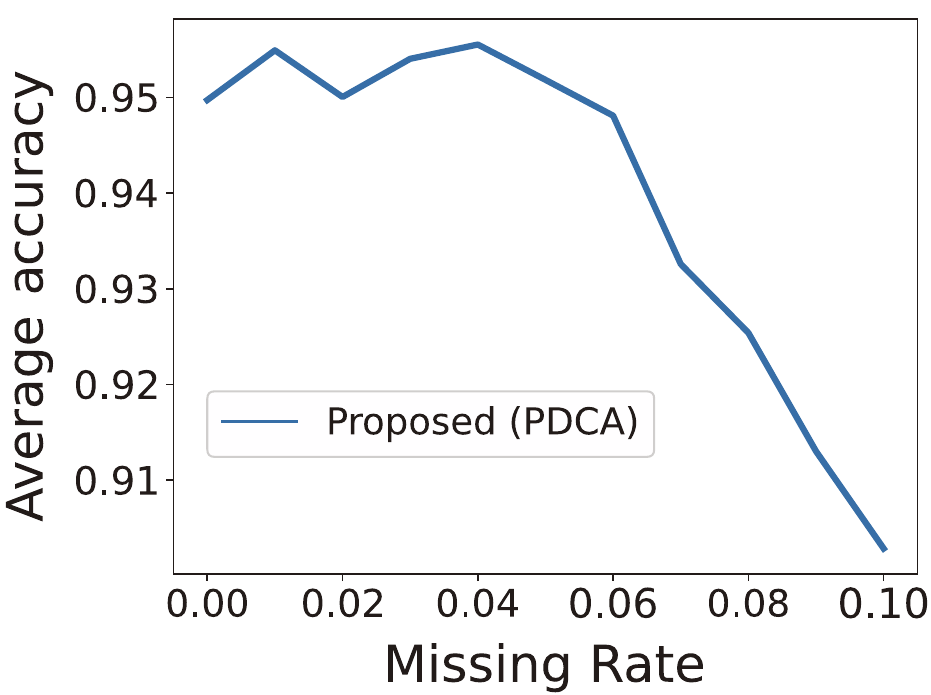}}
    \label{SR}
 \caption{Average accuracy of the proposed method for missing rates from 0 to 0.1 in 0.01 increments ($s=0.1$).}
 \label{fig:missing}
 \end{figure}

\subsection{Experimental results}
Table~\ref{tab:result} shows the image classification accuracy of each method on the two datasets, and Fig.~\ref{fig:plot_round} plots the average accuracies at each communication round in CIFAR-10. 
In Table~\ref{tab:result}, the proposed FL shows the highest accuracy for all noise levels $s$ and datasets. 
Additionally, Fig.~\ref{fig:plot_round} shows that the proposed method exhibits consistently higher accuracy for entire communication rounds than the other methods. 
These results validate that our JGESR aggregation enables efficient training in GFL by effectively estimating the inter-client graph and performing graph-based restoration.

Fig.~\ref{fig:missing} illustrates average accuracies of the proposed method under different missing rates. 
As shown in Fig.~\ref{fig:missing}, our method maintains high average accuracy (i.e., less than $10 \%$ drop) when the missing rates are less than $10\%$. 
This implies the robustness of our method: It can accurately compensate for the missing parameters based on the smoothness assumption.

\section{Conclusion}
In this paper, we propose a robust aggregation method for GFL under noisy communications.
First, we assume that the model parameters, aggregated in the global model update, smoothly vary on the inter-client graph. 
Based on the assumption, we formulate the graph-based aggregation as a DC optimization problem so that it simultaneously optimizes both the graph and model parameters. 
Finally, we solve the DC optimization problem by applying PDCA.
Our method enables efficient training in GFL by effectively estimating the inter-client graph
and performing signal restoration. 
The image classification experiments on two benchmark datasets demonstrate that the proposed FL shows superior prediction performance to the existing methods under noisy conditions.

\bibliographystyle{IEEEbib}
\bibliography{ref1}

\end{document}